\documentclass[letterpaper]{article} 
\usepackage{aaai25}  
\usepackage{times}  
\usepackage{helvet}  
\usepackage{courier}  
\usepackage[hyphens]{url}  
\usepackage{graphicx} 
\usepackage{natbib}  
\usepackage{amsmath}
\usepackage{caption} 
\usepackage{multirow}
\usepackage{ragged2e}
\usepackage{algorithm}
\usepackage{algorithmic}

\usepackage{newfloat}
\usepackage{listings}
\DeclareCaptionStyle{ruled}{labelfont=normalfont,labelsep=colon,strut=off} 
\floatstyle{ruled}
\newfloat{listing}{tb}{lst}{}
\floatname{listing}{Listing}
\title{TSGaussian: Semantic and Depth-Guided Target-Specific Gaussian Splatting from Sparse Views}
\author{
    Liang Zhao\textsuperscript{\rm 1}\equalcontrib,
    Zehan Bao\textsuperscript{\rm 1}\equalcontrib,
    Yi Xie\textsuperscript{\rm 1},
    Hong Chen\textsuperscript{\rm 1,2},
    Yaohui Chen\textsuperscript{\rm 3},
    Weifu Li\textsuperscript{\rm 1,3}\thanks{Corresponding author}\\
}
\affiliations{
    \textsuperscript{\rm 1}College of Informatics, Huazhong Agricultural University, Wuhan, China.\\
    \textsuperscript{\rm 2}Engineering Research Center of Intelligent Technology for Agriculture, Ministry of Education, Wuhan 430070, China.
    \textsuperscript{\rm 3}College of Engineering, Huazhong Agricultural University, Wuhan, China.\\
    liweifu@mail.hzau.edu.cn
}

\usepackage{bibentry}

\begin{document}
\maketitle

\begin{abstract}
Recent advances in Gaussian Splatting have significantly advanced the field, achieving both panoptic and interactive segmentation of 3D scenes. However, existing methodologies often overlook the critical need for reconstructing specified targets with complex structures from sparse views. To address this issue, we introduce TSGaussian, a novel framework that combines semantic constraints with depth priors to avoid geometry degradation in challenging novel view synthesis tasks. Our approach prioritizes computational resources on designated targets while minimizing background allocation. Bounding boxes from YOLOv9 serve as prompts for Segment Anything Model to generate 2D mask predictions, ensuring semantic accuracy and cost efficiency. TSGaussian effectively clusters 3D gaussians by introducing a compact identity encoding for each Gaussian ellipsoid and incorporating 3D spatial consistency regularization. Leveraging these modules, we propose a pruning strategy to effectively reduce redundancy in 3D gaussians. Extensive experiments demonstrate that TSGaussian outperforms state-of-the-art methods on three standard datasets and a new challenging dataset we collected, achieving superior results in novel view synthesis of specific objects. Code is available at: https://github.com/leon2000-ai/TSGaussian.
\end{abstract}

%

\section{Introduction}
Learning 3D representations from 2D images has long been a fundamental objective in computer vision, underpinning a wide range of applications such as augmented reality \cite{Dai_Zhang_Mao_Liu_2020}, robotics \cite{lu2024manigaussian}, and autonomous navigation \cite{jin2024gs}. Recent advances in 3D Gaussian Splatting (3DGS) have improved this field. These advancements allow for more effective reconstruction of 3D scenes \cite{kerbl3Dgaussians}. However, existing methods typically struggle to maintain semantic consistency and avoid geometric degradation when isolating specific targets in cluttered environments \cite{Jain_Tancik_Abbeel_2021,Yu_Ye_Tancik_Kanazawa_2021}. Moreover, optimizing computational resources while preserving the quality of the reconstructed scene remains a significant challenge.

2D semantic segmentation typically requires lower annotation costs compared to 3D methods. Especially, the Segment Anything Model (SAM) has been proven to achieve competitive or even superior zero-shot performance in scene understanding compared to previous supervised models \cite{cen2023segment}. 
As a result, the 2D semantic segmentation of SAM can be employed to enhance the capture of 3D details in 3DGS. For example, GaussianEditors performs semantic tracking based on SAM's semantic segmentation, enabling more precise and efficient editing control \cite{chen2024gaussianeditor}. Gaussian Grouping assigns semantic attributes to each Gaussian primitive based on 2D semantic segmentation \cite{ye2025gaussian}. These methods demonstrate promise in panoramic reconstruction and interactive semantics but face challenges with complex geometries and cluttered scenes \cite{wang2024empirical}. Existing methods often conduct interactive or panoramic segmentation post-reconstruction, resulting in considerable computational redundancy when focusing on specific objects.

\begin{figure*}[!t]
\centering
\includegraphics[scale=1]{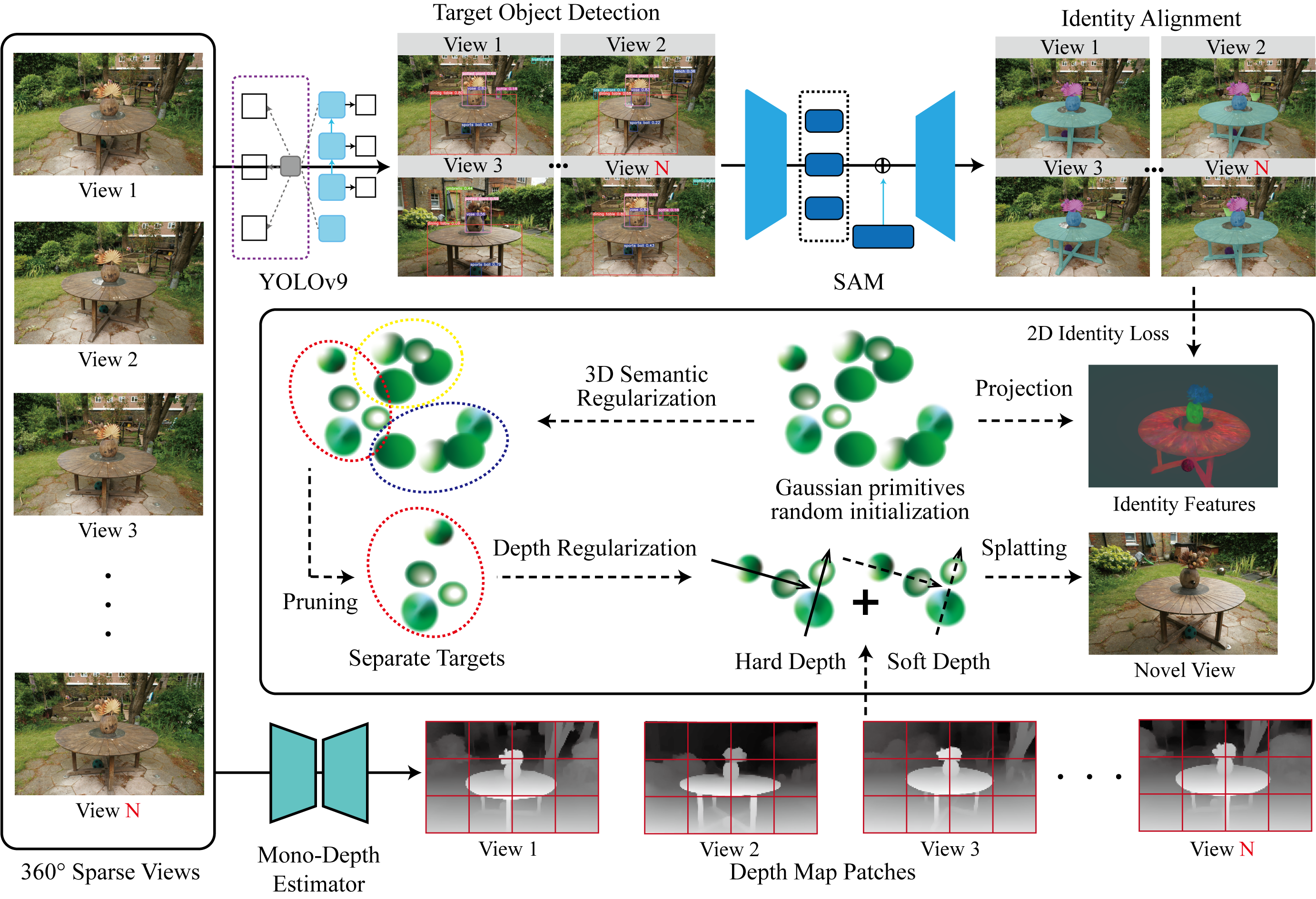}
\caption{Our framework first takes a 360° sparse image sequence as input, using YOLOv9 and SAM to obtain target masks, and a depth estimator to generate depth maps. Next, a general tracking model aligns the identity masks across frames. The framework then randomly generates an initial Gaussians and optimizes the Gaussian field using 2D identity loss, 3D regularization loss, semantic control, and pruning, while performing depth regularization. The final Gaussian field enables depth-accurate and semantically rich target view synthesis.}
\label{fig:TSgaussian}
\end{figure*}

For 3D reconstruction of specific objects, the input image sequence may be relatively sparse due to weather dependency, high capture costs, and time constraints \cite{wang2023sparsenerf}. Although effective for forward-facing scenes in sparse views, they struggle to maintain geometric integrity in omnidirectional reconstruction \cite{niemeyer2022regnerf}. Additionally, few algorithms introduce semantic constraints for reconstructing targeted objects from sparse views. Addressing these limitations requires developing approaches that effectively balance semantic constraints with depth regularization \cite{li2024dngaussian}. To this end, we propose an innovative framework (illustrated in Fig. \ref{fig:TSgaussian}) that facilitates a cost-effective transition from 2D semantic labels to 3D semantic understanding, with a specific emphasis on the rapid reconstruction of targeted objects. Our approach is engineered to ensure robustness even when faced with sparse views. This paper primarily contributes the following:
\begin{itemize}
\item We employ YOLOv9 for scene comprehension, utilizing it as a prompt for SAM to achieve cost-effective 2D semantic segmentation, which guides the training of 3D semantic encoding.
\item We design a semantic operator to identify unnecessary Gaussians during target reconstruction and implement a pruning strategy to minimize redundant computations.
\item We integrate monocular depth estimation as a prior and utilize the depth estimation loss to enhance the robustness of 3DGS in sparse views.
\end{itemize}

\section{Related Work}

\subsection{3D Gaussian Models for Novel View Synthesis}
The advancement of 3DGS has emerged as a crucial technique for novel view synthesis. A growing body of research has introduced significant enhancements to 3DGS, refining its methodologies and broadening its scope of application. To mitigate artifacts during the reconstruction process, Mip-splatting incorporates a smoothing filter that regulates the size of Gaussian primitives by controlling the maximum sampling frequency \cite{yu2024mip}. VDGS further advances the field by proposing using a neural network, similar to NeRF, to replace spherical harmonics in the original 3DGS \cite{malarz2023gaussian}. This innovation improves colour and opacity attributes, yielding more realistic rendering effects. LightGaussian accelerates the rendering speed by identifying Gaussian primitives with minimal contribution to scene reconstruction \cite{fan2023lightgaussian}. It employs a pruning and restoration process that effectively reduces redundancy in Gaussian counts while preserving the visual quality. Despite the significant progress \cite{lu2024scaffold,morgenstern2025compact}, existing methods struggle to achieve high-quality reconstruction for specific semantic targets in complex environments. Therefore, exploring automated 3D reconstruction methods for specific targets remains a major challenge.

\subsection{3D Scene Understanding}
Understanding and tracking the 3D scene are vital to achieve 3D reconstruction for specific targets \cite{takmaz2023openmask3d}. However, the complexity of 3D shapes poses a challenge in maintaining semantic consistency. Mainstream methods typically integrate 2D mask predictions from SAM with 3D spatial consistency constraints to embed semantic features into 3D scene representations \cite{zhang2023faster}, enabling effective segmentation, understanding, and editing of scenes. Geometry-Preserving Neural Radiance Fields (GP-NeRF) further integrates the Transformer architecture to jointly aggregate radiance and semantic embedding information, enhancing the discrimination and quality of the semantic field while maintaining geometric consistency \cite{li2024gp}. However, it is difficult to assign semantic labels to each voxel accurately using static masks in the implicit 3D scene. In contrast, the explicit 3DGS representation method directly assigns semantic labels to each Gaussian primitive, thus optimizing semantic tracking in 3D scenes \cite{chen2024gaussianeditor}. For example, Gaussian Grouping introduces an identity code as a learnable attribute for each Gaussian primitive. This identity code ensures cross-frame consistency and facilitates semantic acquisition \cite{ye2025gaussian}. However, SAM-based interactive methods require frequent manual adjustments to generate 2D semantic masks for specific targets. Additionally, SAM-based panoptic segmentation may fall short in handling complex scenes \cite{kirillov2023segment}, often necessitating the integration of object recognition networks \cite{wang2024yolov9}. Moreover, semantic 3D reconstruction typically requires dense view inputs, which significantly increase acquisition costs and hinder the practical application of the algorithm.

\subsection{Sparse Shot Novel View Synthesis}
Although many algorithms depend on dense views to ensure effectiveness, acquiring such views can be challenging in practical applications due to high costs. To address this issue, current research focuses on incorporating depth priors to guide the 3D reconstruction process \cite{yu2021pixelnerf}. The DNGaussian algorithm enhances sensitivity to subtle depth variations and improves reconstruction results by combining soft and hard depth regularization along with global-local depth normalization \cite{li2024dngaussian}. SparseNeRF, on the other hand, leverages a pre-trained depth estimation model to predict depth maps and introduces local and global depth losses to achieve visually smooth rendering effects \cite{wang2023sparsenerf}. Despite attempts to use diffusion models for novel view generation from sparse views, these methods are inefficient in handling complex scenes and struggle to meet practical application demands. However, the aforementioned studies hardly focus on the reconstruction of specific targets from sparse views \cite{li2023gaussiandiffusion,tang2023dreamgaussian,feng2024fdgaussian}.

\section{Method}
Building on recent advancements in 3DGS, we develop an algorithm that extends high-performing 2D scene understanding techniques to the 3D domain in sparse views. As illustrated in Fig. \ref{fig:TSgaussian}, our approach focuses on constructing a 3D scene representation for specific targets by utilizing semantic and depth constraints. The proposed TSGaussian offers the following technical highlights: 1) 2D detection, semantic segmentation, and 3D reconstruction of specific target objects; 2) Separation of reconstructed 3D objects based on their semantic identities, enabling distinct handling of different semantic components; 3) High-quality 3D reconstruction and rendering in sparse-view scenarios without compromising reconstruction quality.

\subsection{Consistent Targeted Semantic Segmentation}
In order to obtain accurate semantic information, we first integrate YOLOv9 and SAM to generate more accurate 2D semantic masks. Subsequently, a zero-shot tracker is employed to align semantic identities across different views.

\begin{figure}[htbp]
\centering
\includegraphics{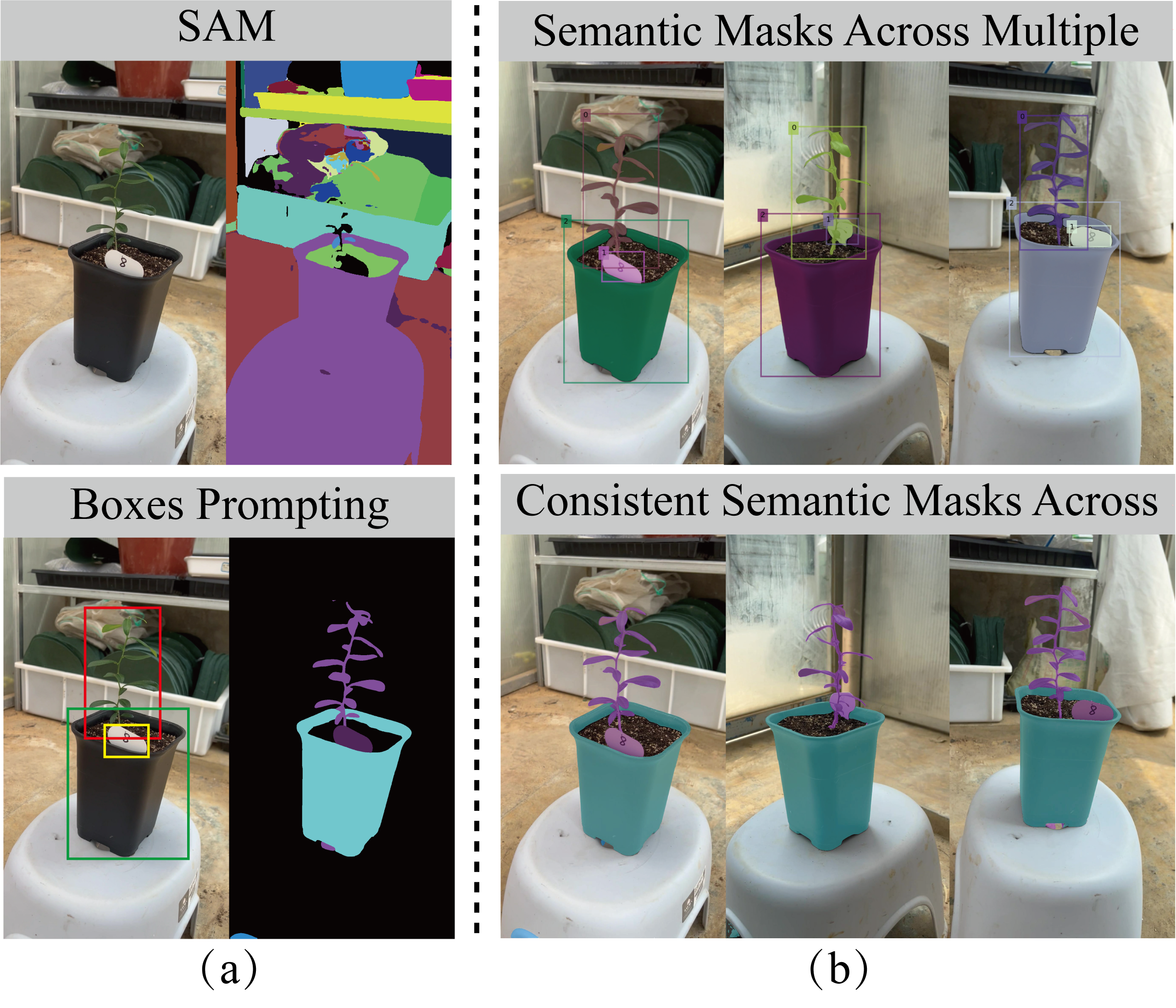} 
\caption{SAM-based panoramic segmentation can recognize common scenes, while SAM with prompts can easily extend to custom scenes and provide more complete masks.}
\label{fig:sam}
\end{figure}

\textbf{2D Mask Acquisition.} 
We deploy the YOLOv9 model with pre-trained weights to identify targets within the multi-view image collection, generating bounding boxes for each image and thus determining the total number of targets in the 3D scene. This method requires only low-cost annotations and fine-tuning during training, making it scalable to complex targets within custom datasets. To obtain the corresponding 2D masks for each target, we use these bounding boxes as prompts for SAM, which automatically generates 2D masks for each image. As shown in Fig. \ref{fig:sam}(a), our approach captures the semantic masks that focus exclusively on the specified targets, which is superior to the coarse and panoramic semantics obtained using SAM alone.

\textbf{Consistent Identity Mapping Across Views.} As shown in Fig. \ref{fig:sam}(b), targets with identical semantics may be assigned different identity documents (IDs) across multiple views. To ensure the consistency, a well-trained zero-shot tracker is employed to correlate the IDs \cite{cheng2023tracking}. This approach helps associate masks with the same identity across different views and assigns a unique ID to each 2D mask within the 3D scene.

\subsection{Semantic Constraints for 2D-to-3D}
To ensure that the results of 2D identity tracking effectively supervise the 3DGS rendering process, Identity Encoding is utilized to assign semantic attributes to the Gaussians. This is followed by a Semantic Rendering module that projects the Gaussians back into 2D semantic masks for calculating the 2D identity loss. Additionally, a 3D semantic regularization loss is introduced to enhance consistency. To further improve the efficiency of the reconstruction process, we implement a Semantic-Driven Gaussians Control and Pruning strategy, which adaptively clones, splits, and prunes the Gaussians based on their semantic attributes.

\textbf{Identity Encoding and Semantic Rendering.} We introduce an identity encoding mechanism for each Gaussian function, where the identity code is a learnable, highly compact vector that assigns a unique instance ID to each target object. To optimize these identity codes, a differentiable Gaussian renderer is employed, allowing the identity code to become a learnable attribute. This approach enables end-to-end training using optimization algorithms like gradient descent. To achieve the semantic rendered masks, we have employed a neural-point-based $\alpha^{\prime}-$rendering technique, where $\alpha^{\prime}$ denotes the influence weight assessed for each pixel \cite{Kopanas_Leimkühler_Rainer_Jambon_Drettakis_2022,Kopanas_Philip_Leimkühler_Drettakis_2021}. The identity feature $E_{\mathrm{id}}$ of the rendered 2D mask for each pixel is obtained by a weighted sum of the identity codes as the following:
\begin{equation}
E_{\text{id}}=\sum_{i\in\mathcal{N}}e_i\alpha_i'\prod_{j=1}^{i-1}(1-\alpha_j'),
\end{equation}
where the $e_i$ is a 16-dimensional identity encoding for each Gaussian. The identity feature $E_{\mathrm{id}}$ can be transformed into identity classification through a linear layer followed by a softmax layer as $softmax(f(E_{\mathrm{id}}))$. For simplicity, this will be denoted as $F(E_{\mathrm{id}})$ in the following text.

\textbf{Grouping Loss.} The grouping loss consists of 2D identity loss and 3D regularization loss. The 2D identity loss $\mathcal{L}_{2\mathrm{d}}$ is calculated as:
\begin{equation}
\mathcal{L}_{2\mathrm{d}}=H(P,F(E_{\mathrm{id}})),
\end{equation}
where $H$ represents the cross-entropy loss and $P$ denotes the correct identity classification. To ensure that the identity encodings of the top $K$-nearest 3D Gaussians are closely matched in terms of their feature distances, the 3D regularization loss for $m$ sampling points is formalized as follows:
\begin{equation}
    \mathcal{L}_{\mathrm{3d}}=\frac{1}{mK}\sum_{j=1}^{m}\sum_{i=1}^{K}F(e_{j})\log\left(\frac{F(e_{j})}{F(e_{i}^{\prime})}\right).
\end{equation}
The grouping loss $\mathcal{L}_{\mathrm{id}}$ is ultimately computed as follows:
\begin{equation}
    \mathcal{L}_{\mathrm{id}} = \lambda_{2\mathrm{d}}\mathcal{L}_{2\mathrm{d}}+\lambda_{3\mathrm{d}}\mathcal{L}_{3\mathrm{d}}.
\end{equation}

\textbf{Semantic-Driven Gaussian Control and Pruning.} The 3DGS technique employs adaptive control to dynamically adjust the Gaussian density, transitioning from a sparse to a denser configuration. However, this process relies solely on view-space position gradients and does not account for semantic constraints, which may result in the generation and accumulation of semantically incorrect Gaussians. To address this issue, we implement a control mechanism based on Gaussian semantic attributes during the densification process. First, we determine the region of interest (ROI) in 3D space according to the semantic attributes of the Gaussians. View-space position gradients are then computed only for Gaussians within this ROI. Gaussians in under-reconstructed areas are cloned, while those in over-reconstructed regions are split. Additionally, since the semantic attributes of Gaussians may change during the densification iterations, we apply pruning to remove Gaussians that no longer belong to the ROI, thereby reducing the accumulation of error. A floating mask is introduced for each training view to leverage the explicit representation of the 3D Gaussian distribution, eliminating incorrect semantic artifacts.

\begin{figure}[!t]
	\centering
	\includegraphics[scale=1]{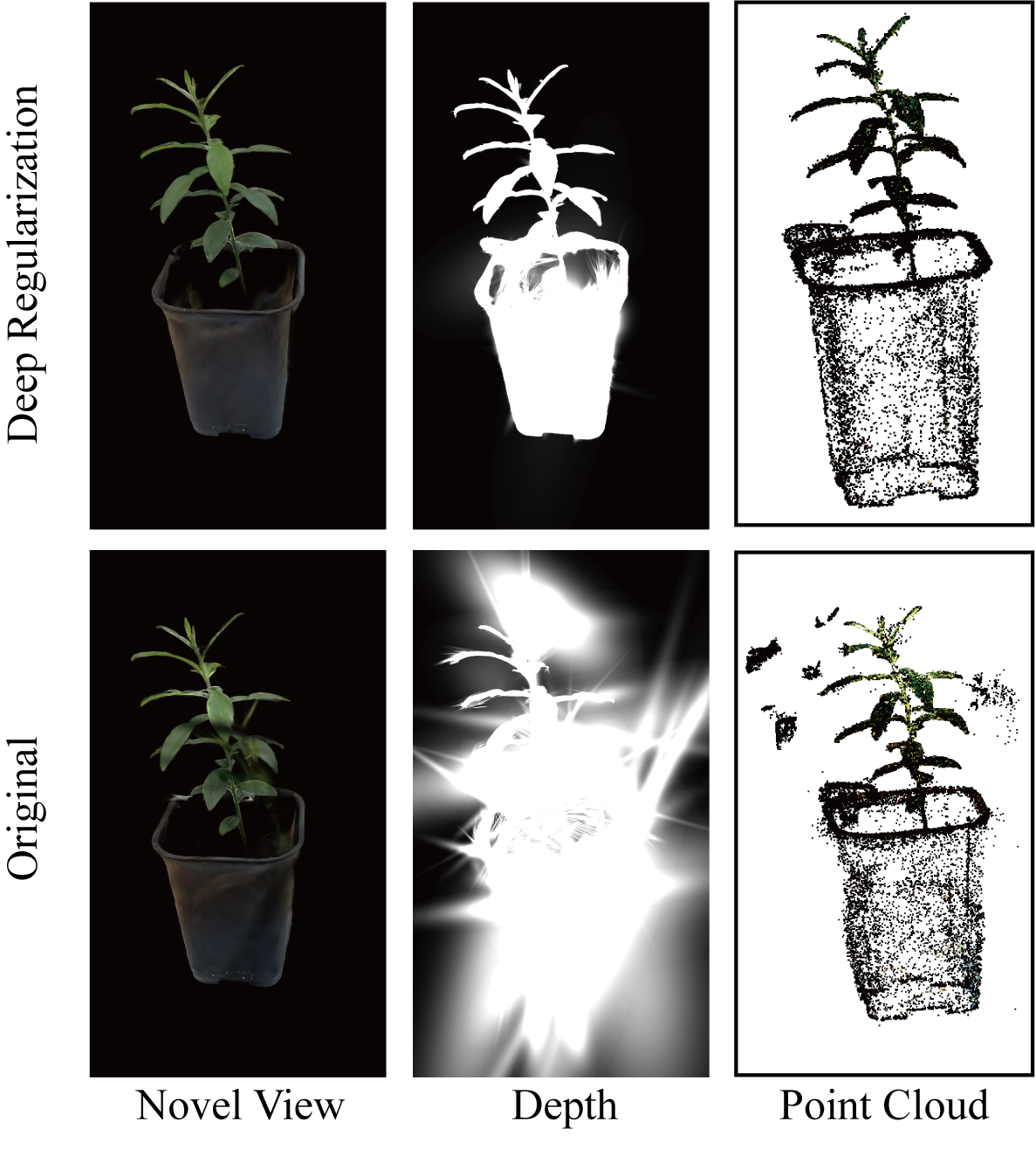} 
	\caption{The training process based on 2D views pays limited attention to depth errors, which can lead to inaccuracies during the training of sparse views. We employ a combination of global and local depth regularization to reduce artifacts, aiding in the acquisition of a model with more precise depth accuracy.}
	\label{fig:depth}
\end{figure}

\subsection{Multi-Scale Depth Regularization}
As illustrated in Fig. \ref{fig:depth}, insufficient attention to depth errors also leads to artifacts in the context of sparse views. To mitigate this issue, we incorporate a monocular depth estimator as an additional spatial geometry prior to generate depth maps for each input view \cite{Ranftl2020}. To avoid over-fitting on the target depth map, a multi-scale depth regularization loss including a soft-hard depth loss and a global-local depth loss is introduced to learn the shape parameters $\{\mu,s,q,\alpha\}$ of 3D Gaussians and enhance sensitivity to depth errors. 
\begin{figure*}[!t]
\centering
\includegraphics[scale=1]{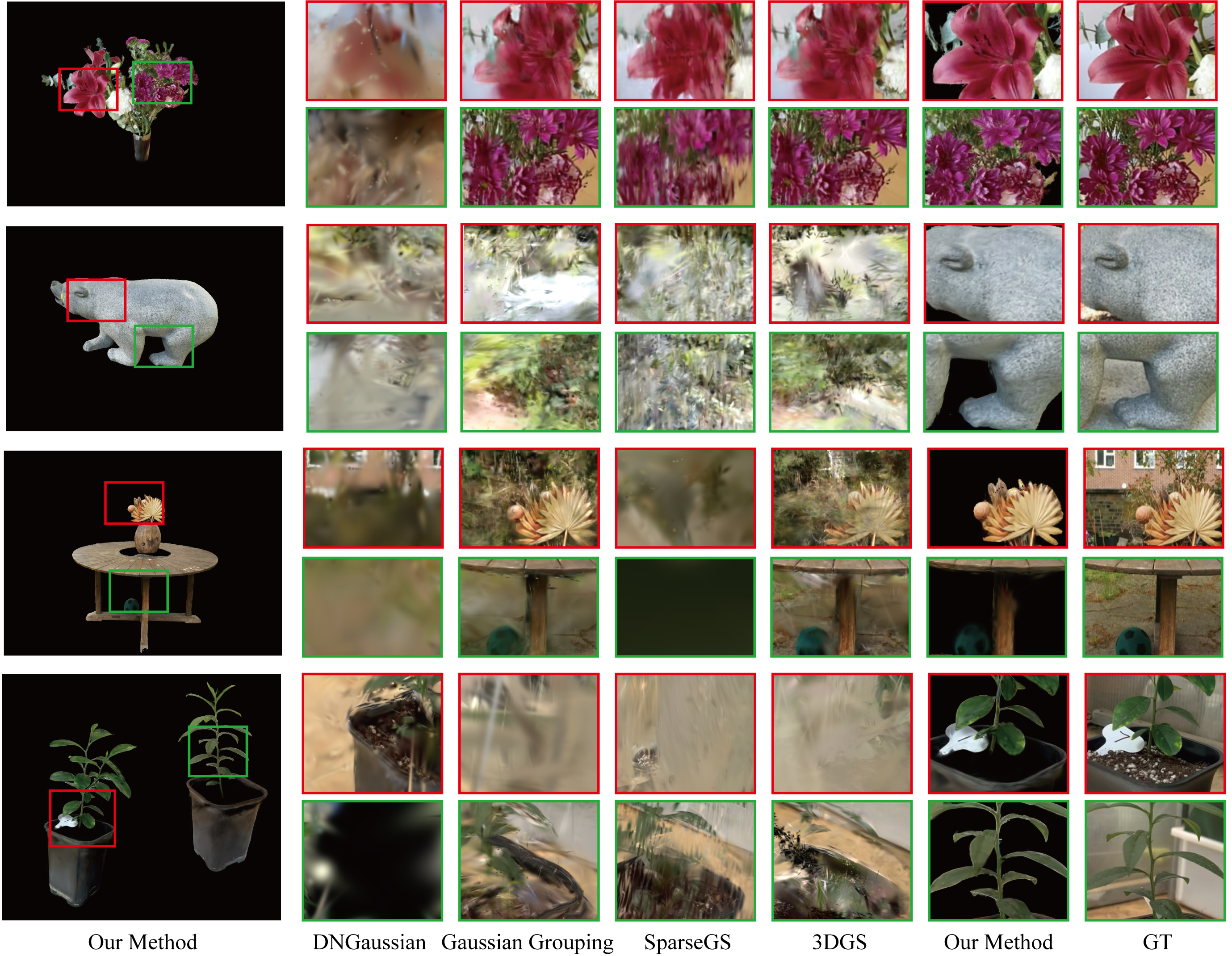}
\caption{Result for 3D reconstruction of specific semantic targets under sparse-view. TSGaussian excels by generating high-quality novel views of specific targets while preserving fine model details.}
\label{fig:result}
\end{figure*}

\textbf{Soft-Hard Depth Loss.}  The soft-hard depth loss specifically focuses on the opacity $\alpha$ and center $\mu$, as these parameters represent the object's spatial occupancy and location, respectively. During the depth regularization process, the scale parameter $s$ and the rotation parameter $q$ are kept fixed to prevent overfitting. A hard depth map $\mathcal{D}_{hard}$ is rendered, primarily composed of the nearest Gaussians along the rays emanating from the camera center and passing through each pixel. In this process, only the center $\mu$ remains as an optimizable parameter, with Gaussian center regularization encouraging the hard depth to align with the monocular depth estimation. The hard depth loss over the target objects $T$ is calculated as:
\begin{equation}
    \mathcal{L}_{hard}(\mathcal{T})=\mathcal{L}_{2}(\mathcal{D}_{hard}(\mathcal{T}),\widetilde{\mathcal{D}}(\mathcal{T})),
\end{equation}
where $\widetilde{\mathcal{D}}$ represents the output of the depth estimator. Similarly, we fix the Gaussian center $\mu$, render a soft depth map $\mathcal{D}_{soft}$, and use depth regularization to adjust the opacity $\alpha$. The soft depth loss for this process is as follows:
\begin{equation}
    \mathcal{L}_{soft}(\mathcal{T})=\mathcal{L}_{2}(\mathcal{D}_{soft}(\mathcal{T}),\widetilde{\mathcal{D}}(\mathcal{T})).
\end{equation}
Then the soft-hard depth loss is formulated by:
\begin{equation}
\mathcal{L}_{\mathrm{SH}} = \mathcal{R}_{hard} + \mathcal{R}_{soft}.
\end{equation}

\textbf{Global-Local Depth Loss.} The global-local depth loss is employed to finely correct minor errors in depth estimation. To achieve this, the predicted depth map and the depth estimator's output are divided into smaller patches, which are then normalized on a local scale to have a mean value of 0 and a standard deviation close to 1. The normalized result is denoted as $\mathcal{D}^{LN}$. We utilize the global standard deviation of the depth map in place of the standard deviation of the local blocks to obtain $\mathcal{D}^{GN}$. Then the global-local depth loss $\mathcal{L}_{GL}$ is defined as:
\begin{equation}
    \mathcal{L}_{GL}=\mathcal{L}_{2}(\mathcal{D}_{T}^{GN},\widetilde{\mathcal{D}}^{GN})+\gamma\mathcal{L}_{2}(\mathcal{D}_{T}^{LN},\widetilde{\mathcal{D}}^{LN}).
\end{equation}
Thus, the multi-scale depth loss is formulated by:
\begin{equation}
    \mathcal{L}_{\mathrm{D}} = \lambda_{\mathrm{SH}}\mathcal{L}_{\mathrm{SH}}+\lambda_{\mathrm{GL}}\mathcal{L}_{GL}.
\end{equation}

For color reconstruction, we combine an L1 reconstruction loss with a D-SSIM measure to ensure the structural similarity between the rendered image and the actual image:
\begin{equation}
\mathcal{L}_{color}=\mathcal{L}_1(\hat{\mathcal{I}},\mathcal{I})+\lambda\mathcal{L}_{\mathrm{D-SSIM}}(\hat{\mathcal{I}},\mathcal{I}).
\end{equation}
The total loss function for training is formulated by:
\begin{equation}
    \mathcal{L} = \mathcal{L}_{color}+\lambda_{\mathrm{id}}\mathcal{L}_{\mathrm{id}}+\lambda_{\mathrm{D}}\mathcal{L}_{\mathrm{D}}.
\end{equation}


\begin{table*}[htbp]
  \renewcommand{\arraystretch}{1.15}
  \centering
  
    \begin{tabular}{ccccc}
    \hline
    \multirow{2}{*}{Method} & \textit{\textbf{``bear"}} & \textit{\textbf{``bouquet"}} & \textit{\textbf{``garden"}} & \textit{\textbf{Ours dataset}} \\
          & PSNR/ SSIM/ LPIPS & PSNR/ SSIM/ LPIPS  & PSNR/ SSIM/ LPIPS  & PSNR/ SSIM/ LPIPS  \\ \hline
    DNGaussian & 19.53/0.850/0.141 & 17.14/0.818/0.177 & 20.97/0.900/0.795 & 17.90/0.889/0.109 \\
    Gaussian Grouping & 21.97/0.873/0.104 & 17.99/0.837/0.127 & 24.27/0.937/0.053 & 18.90/0.921/0.078 \\
    SparseGS  & 20.75/0.852/0.119 & 17.26/0.819/0.819 & 22.16/0.502/0.502 & 18.16/0.909/0.092 \\
    3DGS  & 20.71/0.861/0.115 & 17.05/0.822/0.150 & 26.05/0.945/0.042 & 18.35/0.917/0.081 \\
    TSGaussian (Ours) & 26.99/0.894/0.082 & 20.80/0.942/0.128 & 27.93/0.942/0.049 & 27.40/0.942/0.063 \\ \hline
    \end{tabular}%
  
  \caption{A quantitative comparison between sparse input views on public dataset scenes and our self-collected data scenes. It is important to note that under 360° sparse view conditions, the model is prone to overfitting on the training views, resulting in significant artifacts on unseen views, which leads to lower evaluation metrics.}
  \label{tab:result}%
\end{table*}%

\section{Experiments and Analysis}

\subsection{Dataset and Experiment Setup}

\textbf{Datasets.} Unlike panoramic reconstruction, this research focuses on scenes dedicated to specific targets for the evaluation. Notable scenes include ``garden" from Mip-NeRF 360 \cite{barron2022mip}, ``bouquet" from LERF Datasets \cite{kerr2023lerf}, and ``bear" from Gaussian Grouping \cite{ye2025gaussian}. Additionally, we employ smartphone cameras to perform 360° panoramic imaging of targeted Citrus, which is crucial for accurately extracting their phenotypic traits. This dataset supports a more precise evaluation of 3D reconstruction methods applied to specific objects.

\begin{figure}[!t]
\centering
\includegraphics[scale=1]{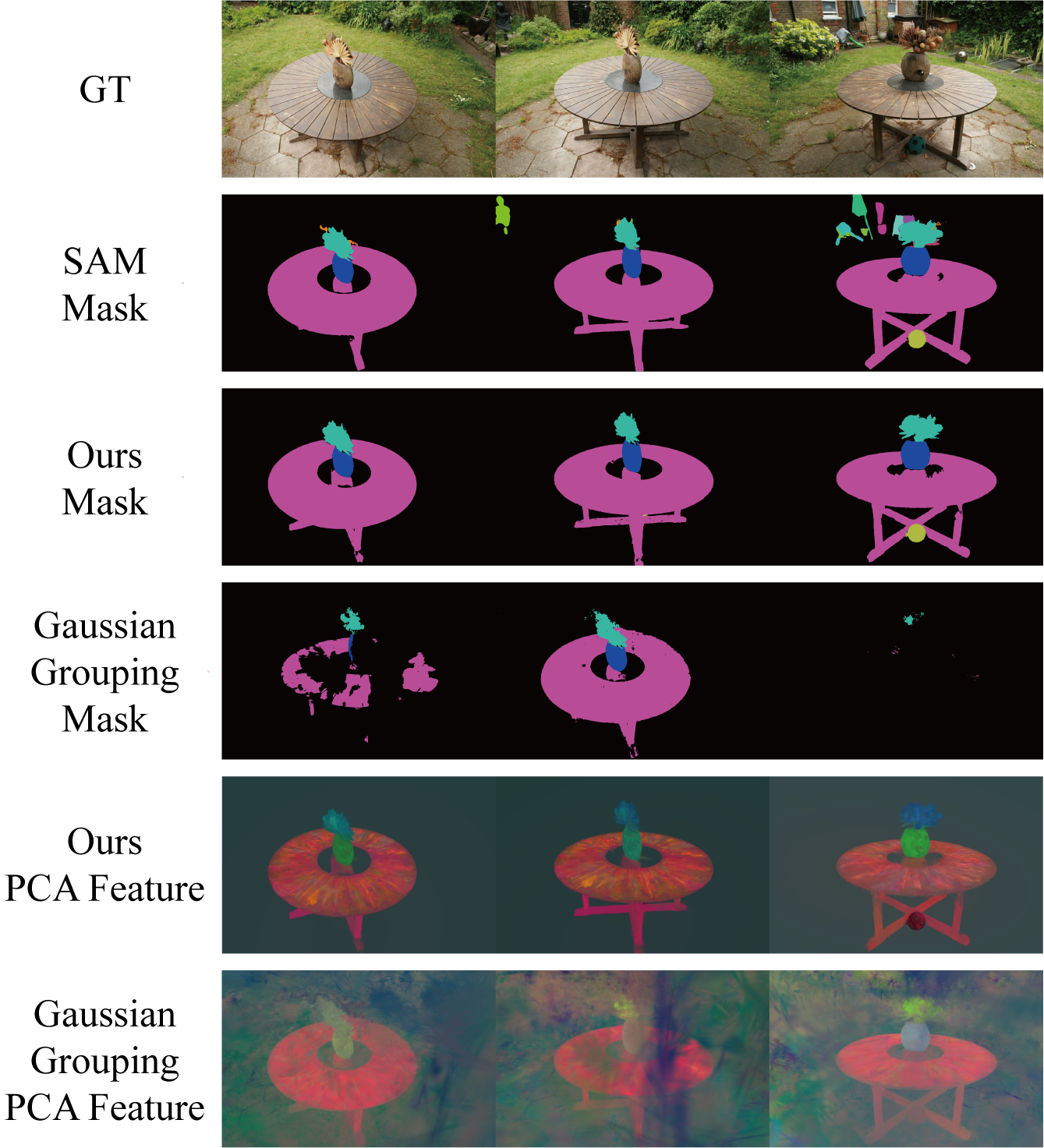} 
\caption{A comparison of view segmentation between Gaussian Grouping and our proposed method in rendering. The masks predicted by Gaussian Grouping exhibit significant errors due to geometric degradation caused by sparse views, resulting in occlusion by artifacts. In contrast, our method, enhanced by semantic constraints and depth regularization, substantially reduces these artifacts. The identity encoding features in the bottom row are visualized using Principal Component Analysis (PCA).}
\label{fig:mask}
\end{figure}

\textbf{Experimental Settings.} We refer to the settings in previous studies to sample different scenes and form sparse views. Since our research goal is to reconstruct a 360° scene of a specific object, we uniformly extracted one-third of the original views to ensure completeness and divided these views evenly into training and test sets. Inspired by previous sparse view setting, the camera poses are assumed to be known through calibration or other methods. The number of views is set as 10 in the  ``bear" scene, and is 30 for the larger ``bouquet" and ``garden" scenes. Note that the output bounding boxes of YOLOv9 model serve as the prompts for the SAM to generate high-quality masks. The pretrained YOLOv9 on the COCO dataset is directly utilized to detect targets for the public scenes. 
In our own dataset, a low-cost box annotation method using 12 videos is employed to fine-tune the pretrained YOLOv9 model, thereby enhancing its ability to detect citrus targets. Furthermore, to address the recognition challenges of YOLOv9 in target frames, we also use the output bounding boxes from adjacent frames as the detection output, ensuring the integrity of the mask.

\textbf{Baselines.} We compare the proposed method with the original 3D GS \cite{kerbl3Dgaussians}, as well as its variants including Gaussian Grouping \cite{ye2025gaussian}, DNGaussian \cite{li2024dngaussian}, and SparseGS \cite{xiong2023sparsegs}. To ensure a fair comparison, all compared algorithms use the same semantic masks of specific objects.

\textbf{Implementation Details.} The naive COLMAP is utilized to obtain the camera poses \cite{schonberger2016structure}. The semantic masks are employed based on SAM with prompts by YOLOv9 and tracking was performed with DEVA to ensure cross-view identity consistency for each scene \cite{cheng2023tracking}.
We randomly initialize 10,000 points as the initial gaussian. The model was trained for 10,000 iterations using an NVIDIA A6000 GPU.

\subsection{Comparison Results}
 
\textbf{Public Dataset.} The quantitative and qualitative analyses are shown in Table \ref{tab:result} and Fig. \ref{fig:result}, respectively. We found that other models often suffer from overfitting when reconstructing specific targets from sparse views. On public dataset, we outperform all baselines in terms of SSIM, LPIPS, and PSNR. In the ``bear" scene, our PSNR exceeds that of 3DGS by 6.28. In the ``bouquet" scene, our SSIM is 0.12 higher than that of 3DGS. Across various scenes, our method consistently achieves lower LPIPS values. 
In qualitative analysis, our method consistently produces clear outputs across all scenes and accurately recovers the geometric structure. A key feature of our approach is its ability to leverage depth priors under semantic constraints, which avoids overfitting and enhances generalization to new viewpoints under sparse view conditions. In Fig. \ref{fig:mask}, our semantic rendering results are more accurate with fewer artifact interferences, and more accurate Gaussian primitive semantics can be observed in the PCA feature image.

\textbf{Our Dataset.} 
The dataset we collected primarily focuses on Citrus with varying shapes. These plant seedlings are quite slender, posing a significant challenge in sparse-view 3D reconstruction. As shown in Fig. \ref{fig:result}, most methods fail on this dataset due to overfitting. However, our approach consistently maintains robust geometric structures and delivers finely detailed modeling results on various semantic objects such as plants and flower pots. In quantitative analysis, our method's PSNR exceeds that of the best-performing Gaussian Grouping by 8.50, and our SSIM is 0.021 higher.

\subsection{Ablation and Analysis}

\begin{figure}[!t]
\centering
\includegraphics[scale=1]{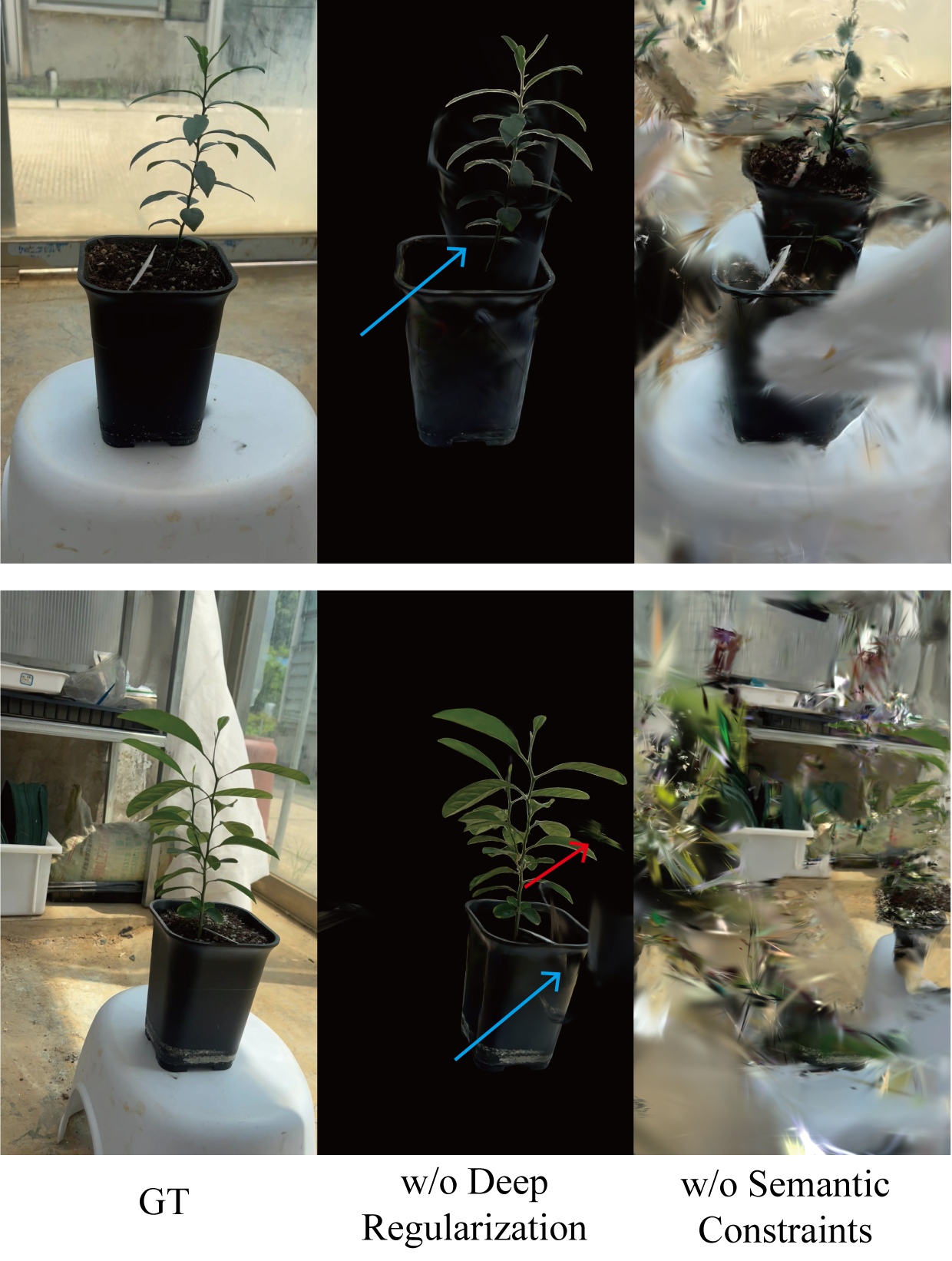} 
\caption{Visualization results of ablation study. In the synthetic new views without depth regularization, artifacts are visible at the locations indicated by the blue and red arrows. While these artifacts may not significantly affect the evaluation metrics from certain angles, they impact the distribution of Gaussian primitives. Without semantic constraints, Gaussian primitives are distributed globally rather than being concentrated on the target, which can easily lead to overfitting on the training views.}
	\label{fig:ablation}
\end{figure}

\textbf{Ablation of Deep Regularization.}   
As indicated by the red arrow in Fig. \ref{fig:ablation}, the absence of depth regularization will lead to artifacts in the background, which affect the Gaussian representation of specific targets. Although these artifacts may not be prominent from certain rendering angles, they can still yield favorable results in quantitative analysis, as validated in Table \ref{tab:ablation}, the absence of depth regularization resulted in a PSNR decrease of 0.496 and an SSIM decrease of 0.014.
 
\textbf{Ablation of Semantic Constraints.} To accurately assess the impact of semantic constraints, we provide the rendering results without semantic constraints in Fig. \ref{fig:ablation} and Table \ref{tab:ablation}. We observed that the absence of semantic constraints leads to redundant background information, which significantly degrades performance and results in a PSNR decrease of 9.349 and an SSIM decrease of 0.03. The experiment highlights the importance of the semantic information obtained through object detection and SAM for target reconstruction from sparse views.

\section{Conclusion}

In this study, we have proposed a novel approach that integrates depth regularization and semantic constraints to enhance the performance of 3D Gaussian Splatting. To minimize the impact of background noise, we optimize the Gaussian distribution by leveraging the SAM-based semantic segmentation with prompts from YOLOv9. Despite potential inaccuracies in 2D semantic information, our framework can achieve robust recognition in complex environments by utilizing the consistency across 3D views. Furthermore, we have introduced Multi-Scale Depth Regularization to reduce data acquisition costs and minimize redundant information, effectively mitigating artifacts during the reconstruction process. This method proves effective for studying target objects from sparse views, with the pruning of Gaussian primitives for specific targets. The experiments highlight the importance of integrating semantic and depth information in 3D reconstruction tasks, paving the way for future advancements in this field and expanding the applicability of our approach across diverse scenarios.

\begin{table}[!t]
\begin{tabular}{l|lll}
\hline
Setting                  & PSNR↑  & SSIM↑ & LPIPS↓ \\ \hline
w/o Deep Regularization  & 26.903 & 0.927 & 0.073  \\
w/o Semantic Constraints & 18.046 & 0.912 & 0.086  \\
ALL                      & 27.395 & 0.942 & 0.062  \\ \hline
\end{tabular}
\caption{Ablation study on depth and semantic constraints.}
\label{tab:ablation}%
\end{table}

\section{Limitations and Future Work}
We strongly believe that 3DGS with semantic constraints and depth regularization holds significant potential for enhancing the quality of target-specific reconstruction and minimizing artifacts. By focusing on the precise reconstruction of specific targets, this approach not only establishes a robust foundation for subsequent 3D model applications but also contributes to the broader advancement of the field. Our next goal is to improve the combination of semantic and depth data to make the Gaussian reconstruction method more efficient and effective.




\bibliography{aaai25}

\begin{thebibliography}{35}
\providecommand{\natexlab}[1]{#1}

\bibitem[{Barron et~al.(2022)Barron, Mildenhall, Verbin, Srinivasan, and Hedman}]{barron2022mip}
Barron, J.~T.; Mildenhall, B.; Verbin, D.; Srinivasan, P.~P.; and Hedman, P. 2022.
\newblock Mip-nerf 360: Unbounded anti-aliased neural radiance fields.
\newblock In \emph{Proceedings of the IEEE/CVF conference on computer vision and pattern recognition}, 5470--5479.

\bibitem[{Cen et~al.(2023)Cen, Zhou, Fang, Shen, Xie, Jiang, Zhang, Tian et~al.}]{cen2023segment}
Cen, J.; Zhou, Z.; Fang, J.; Shen, W.; Xie, L.; Jiang, D.; Zhang, X.; Tian, Q.; et~al. 2023.
\newblock Segment anything in 3d with nerfs.
\newblock \emph{Advances in Neural Information Processing Systems}, 36: 25971--25990.

\bibitem[{Chen et~al.(2024)Chen, Chen, Zhang, Wang, Yang, Wang, Cai, Yang, Liu, and Lin}]{chen2024gaussianeditor}
Chen, Y.; Chen, Z.; Zhang, C.; Wang, F.; Yang, X.; Wang, Y.; Cai, Z.; Yang, L.; Liu, H.; and Lin, G. 2024.
\newblock Gaussianeditor: Swift and controllable 3d editing with gaussian splatting.
\newblock In \emph{Proceedings of the IEEE/CVF Conference on Computer Vision and Pattern Recognition}, 21476--21485.

\bibitem[{Cheng et~al.(2023)Cheng, Oh, Price, Schwing, and Lee}]{cheng2023tracking}
Cheng, H.~K.; Oh, S.~W.; Price, B.; Schwing, A.; and Lee, J.-Y. 2023.
\newblock Tracking anything with decoupled video segmentation.
\newblock In \emph{Proceedings of the IEEE/CVF International Conference on Computer Vision}, 1316--1326.

\bibitem[{Dai et~al.(2020)Dai, Zhang, Mao, and Liu}]{Dai_Zhang_Mao_Liu_2020}
Dai, J.; Zhang, Z.; Mao, S.; and Liu, D. 2020.
\newblock A View Synthesis-Based 360° VR Caching System Over MEC-Enabled C-RAN.
\newblock \emph{IEEE Transactions on Circuits and Systems for Video Technology}, 3843–3855.

\bibitem[{Fan et~al.(2023)Fan, Wang, Wen, Zhu, Xu, and Wang}]{fan2023lightgaussian}
Fan, Z.; Wang, K.; Wen, K.; Zhu, Z.; Xu, D.; and Wang, Z. 2023.
\newblock Lightgaussian: Unbounded 3d gaussian compression with 15x reduction and 200+ fps.
\newblock \emph{arXiv preprint arXiv:2311.17245}.

\bibitem[{Feng et~al.(2024)Feng, Xing, Wu, and Jiang}]{feng2024fdgaussian}
Feng, Q.; Xing, Z.; Wu, Z.; and Jiang, Y.-G. 2024.
\newblock Fdgaussian: Fast gaussian splatting from single image via geometric-aware diffusion model.
\newblock \emph{arXiv preprint arXiv:2403.10242}.

\bibitem[{Jain, Tancik, and Abbeel(2021)}]{Jain_Tancik_Abbeel_2021}
Jain, A.; Tancik, M.; and Abbeel, P. 2021.
\newblock Putting NeRF on a Diet: Semantically Consistent Few-Shot View Synthesis.
\newblock In \emph{2021 IEEE/CVF International Conference on Computer Vision (ICCV)}.

\bibitem[{Jin et~al.(2024)Jin, Gao, Lu, and Gao}]{jin2024gs}
Jin, R.; Gao, Y.; Lu, H.; and Gao, F. 2024.
\newblock GS-Planner: A Gaussian-Splatting-based Planning Framework for Active High-Fidelity Reconstruction.
\newblock \emph{arXiv preprint arXiv:2405.10142}.

\bibitem[{Kerbl et~al.(2023)Kerbl, Kopanas, Leimk{\"u}hler, and Drettakis}]{kerbl3Dgaussians}
Kerbl, B.; Kopanas, G.; Leimk{\"u}hler, T.; and Drettakis, G. 2023.
\newblock 3D Gaussian Splatting for Real-Time Radiance Field Rendering.
\newblock \emph{ACM Transactions on Graphics}, 42(4).

\bibitem[{Kerr et~al.(2023)Kerr, Kim, Goldberg, Kanazawa, and Tancik}]{kerr2023lerf}
Kerr, J.; Kim, C.~M.; Goldberg, K.; Kanazawa, A.; and Tancik, M. 2023.
\newblock Lerf: Language embedded radiance fields.
\newblock In \emph{Proceedings of the IEEE/CVF International Conference on Computer Vision}, 19729--19739.

\bibitem[{Kirillov et~al.(2023)Kirillov, Mintun, Ravi, Mao, Rolland, Gustafson, Xiao, Whitehead, Berg, Lo et~al.}]{kirillov2023segment}
Kirillov, A.; Mintun, E.; Ravi, N.; Mao, H.; Rolland, C.; Gustafson, L.; Xiao, T.; Whitehead, S.; Berg, A.~C.; Lo, W.-Y.; et~al. 2023.
\newblock Segment anything.
\newblock In \emph{Proceedings of the IEEE/CVF International Conference on Computer Vision}, 4015--4026.

\bibitem[{Kopanas et~al.(2022)Kopanas, Leimkühler, Rainer, Jambon, and Drettakis}]{Kopanas_Leimkühler_Rainer_Jambon_Drettakis_2022}
Kopanas, G.; Leimkühler, T.; Rainer, G.; Jambon, C.; and Drettakis, G. 2022.
\newblock Neural Point Catacaustics for Novel-View Synthesis of Reflections.
\newblock \emph{ACM Transactions on Graphics}, 1–15.

\bibitem[{Kopanas et~al.(2021)Kopanas, Philip, Leimkühler, and Drettakis}]{Kopanas_Philip_Leimkühler_Drettakis_2021}
Kopanas, G.; Philip, J.; Leimkühler, T.; and Drettakis, G. 2021.
\newblock Point‐Based Neural Rendering with Per‐View Optimization.
\newblock \emph{Computer Graphics Forum}, 29–43.

\bibitem[{Li et~al.(2024{\natexlab{a}})Li, Zhang, Dai, Liu, Cheng, Li, Wang, and Han}]{li2024gp}
Li, H.; Zhang, D.; Dai, Y.; Liu, N.; Cheng, L.; Li, J.; Wang, J.; and Han, J. 2024{\natexlab{a}}.
\newblock GP-NeRF: Generalized Perception NeRF for Context-Aware 3D Scene Understanding.
\newblock In \emph{Proceedings of the IEEE/CVF Conference on Computer Vision and Pattern Recognition}, 21708--21718.

\bibitem[{Li et~al.(2024{\natexlab{b}})Li, Zhang, Bai, Zheng, Ning, Zhou, and Gu}]{li2024dngaussian}
Li, J.; Zhang, J.; Bai, X.; Zheng, J.; Ning, X.; Zhou, J.; and Gu, L. 2024{\natexlab{b}}.
\newblock Dngaussian: Optimizing sparse-view 3d gaussian radiance fields with global-local depth normalization.
\newblock In \emph{Proceedings of the IEEE/CVF Conference on Computer Vision and Pattern Recognition}, 20775--20785.

\bibitem[{Li, Wang, and Tseng(2023)}]{li2023gaussiandiffusion}
Li, X.; Wang, H.; and Tseng, K.-K. 2023.
\newblock Gaussiandiffusion: 3d gaussian splatting for denoising diffusion probabilistic models with structured noise.
\newblock \emph{arXiv preprint arXiv:2311.11221}.

\bibitem[{Lu et~al.(2024{\natexlab{a}})Lu, Zhang, Wang, Liu, Lu, and Tang}]{lu2024manigaussian}
Lu, G.; Zhang, S.; Wang, Z.; Liu, C.; Lu, J.; and Tang, Y. 2024{\natexlab{a}}.
\newblock Manigaussian: Dynamic gaussian splatting for multi-task robotic manipulation.
\newblock \emph{arXiv preprint arXiv:2403.08321}.

\bibitem[{Lu et~al.(2024{\natexlab{b}})Lu, Yu, Xu, Xiangli, Wang, Lin, and Dai}]{lu2024scaffold}
Lu, T.; Yu, M.; Xu, L.; Xiangli, Y.; Wang, L.; Lin, D.; and Dai, B. 2024{\natexlab{b}}.
\newblock Scaffold-gs: Structured 3d gaussians for view-adaptive rendering.
\newblock In \emph{Proceedings of the IEEE/CVF Conference on Computer Vision and Pattern Recognition}, 20654--20664.

\bibitem[{Malarz et~al.(2023)Malarz, Smolak, Tabor, Tadeja, and Spurek}]{malarz2023gaussian}
Malarz, D.; Smolak, W.; Tabor, J.; Tadeja, S.; and Spurek, P. 2023.
\newblock Gaussian splatting with nerf-based color and opacity.
\newblock \emph{arXiv preprint arXiv:2312.13729}.

\bibitem[{Morgenstern et~al.(2025)Morgenstern, Barthel, Hilsmann, and Eisert}]{morgenstern2025compact}
Morgenstern, W.; Barthel, F.; Hilsmann, A.; and Eisert, P. 2025.
\newblock Compact 3d scene representation via self-organizing gaussian grids.
\newblock In \emph{European Conference on Computer Vision}, 18--34. Springer.

\bibitem[{Niemeyer et~al.(2022)Niemeyer, Barron, Mildenhall, Sajjadi, Geiger, and Radwan}]{niemeyer2022regnerf}
Niemeyer, M.; Barron, J.~T.; Mildenhall, B.; Sajjadi, M.~S.; Geiger, A.; and Radwan, N. 2022.
\newblock Regnerf: Regularizing neural radiance fields for view synthesis from sparse inputs.
\newblock In \emph{Proceedings of the IEEE/CVF Conference on Computer Vision and Pattern Recognition}, 5480--5490.

\bibitem[{Ranftl et~al.(2020)Ranftl, Lasinger, Hafner, Schindler, and Koltun}]{Ranftl2020}
Ranftl, R.; Lasinger, K.; Hafner, D.; Schindler, K.; and Koltun, V. 2020.
\newblock Towards Robust Monocular Depth Estimation: Mixing Datasets for Zero-shot Cross-dataset Transfer.
\newblock \emph{IEEE Transactions on Pattern Analysis and Machine Intelligence (TPAMI)}.

\bibitem[{Schonberger and Frahm(2016)}]{schonberger2016structure}
Schonberger, J.~L.; and Frahm, J.-M. 2016.
\newblock Structure-from-motion revisited.
\newblock In \emph{Proceedings of the IEEE conference on computer vision and pattern recognition}, 4104--4113.

\bibitem[{Takmaz et~al.(2023)Takmaz, Fedele, Sumner, Pollefeys, Tombari, and Engelmann}]{takmaz2023openmask3d}
Takmaz, A.; Fedele, E.; Sumner, R.~W.; Pollefeys, M.; Tombari, F.; and Engelmann, F. 2023.
\newblock Openmask3d: Open-vocabulary 3d instance segmentation.
\newblock \emph{arXiv preprint arXiv:2306.13631}.

\bibitem[{Tang et~al.(2023)Tang, Ren, Zhou, Liu, and Zeng}]{tang2023dreamgaussian}
Tang, J.; Ren, J.; Zhou, H.; Liu, Z.; and Zeng, G. 2023.
\newblock Dreamgaussian: Generative gaussian splatting for efficient 3d content creation.
\newblock \emph{arXiv preprint arXiv:2309.16653}.

\bibitem[{Wang, Yeh, and Liao(2024)}]{wang2024yolov9}
Wang, C.-Y.; Yeh, I.-H.; and Liao, H.-Y.~M. 2024.
\newblock Yolov9: Learning what you want to learn using programmable gradient information.
\newblock \emph{arXiv preprint arXiv:2402.13616}.

\bibitem[{Wang et~al.(2023)Wang, Chen, Loy, and Liu}]{wang2023sparsenerf}
Wang, G.; Chen, Z.; Loy, C.~C.; and Liu, Z. 2023.
\newblock Sparsenerf: Distilling depth ranking for few-shot novel view synthesis.
\newblock In \emph{Proceedings of the IEEE/CVF International Conference on Computer Vision}, 9065--9076.

\bibitem[{Wang, Zhao, and Petzold(2024)}]{wang2024empirical}
Wang, Y.; Zhao, Y.; and Petzold, L. 2024.
\newblock An empirical study on the robustness of the segment anything model (sam).
\newblock \emph{Pattern Recognition}, 110685.

\bibitem[{Xiong et~al.(2023)Xiong, Muttukuru, Upadhyay, Chari, and Kadambi}]{xiong2023sparsegs}
Xiong, H.; Muttukuru, S.; Upadhyay, R.; Chari, P.; and Kadambi, A. 2023.
\newblock Sparsegs: Real-time 360 $\{$$\backslash$deg$\}$ sparse view synthesis using gaussian splatting.
\newblock \emph{arXiv preprint arXiv:2312.00206}.

\bibitem[{Ye et~al.(2025)Ye, Danelljan, Yu, and Ke}]{ye2025gaussian}
Ye, M.; Danelljan, M.; Yu, F.; and Ke, L. 2025.
\newblock Gaussian grouping: Segment and edit anything in 3d scenes.
\newblock In \emph{European Conference on Computer Vision}, 162--179. Springer.

\bibitem[{Yu et~al.(2021{\natexlab{a}})Yu, Ye, Tancik, and Kanazawa}]{Yu_Ye_Tancik_Kanazawa_2021}
Yu, A.; Ye, V.; Tancik, M.; and Kanazawa, A. 2021{\natexlab{a}}.
\newblock pixelNeRF: Neural Radiance Fields from One or Few Images.
\newblock In \emph{2021 IEEE/CVF Conference on Computer Vision and Pattern Recognition (CVPR)}.

\bibitem[{Yu et~al.(2021{\natexlab{b}})Yu, Ye, Tancik, and Kanazawa}]{yu2021pixelnerf}
Yu, A.; Ye, V.; Tancik, M.; and Kanazawa, A. 2021{\natexlab{b}}.
\newblock pixelnerf: Neural radiance fields from one or few images.
\newblock In \emph{Proceedings of the IEEE/CVF conference on computer vision and pattern recognition}, 4578--4587.

\bibitem[{Yu et~al.(2024)Yu, Chen, Huang, Sattler, and Geiger}]{yu2024mip}
Yu, Z.; Chen, A.; Huang, B.; Sattler, T.; and Geiger, A. 2024.
\newblock Mip-splatting: Alias-free 3d gaussian splatting.
\newblock In \emph{Proceedings of the IEEE/CVF Conference on Computer Vision and Pattern Recognition}, 19447--19456.

\bibitem[{Zhang et~al.(2023)Zhang, Han, Qiao, Kim, Bae, Lee, and Hong}]{zhang2023faster}
Zhang, C.; Han, D.; Qiao, Y.; Kim, J.~U.; Bae, S.-H.; Lee, S.; and Hong, C.~S. 2023.
\newblock Faster segment anything: Towards lightweight sam for mobile applications.
\newblock \emph{arXiv preprint arXiv:2306.14289}.

\end{thebibliography}

\end{document}